\documentclass{article}
\usepackage{iclr2026_conference,times}
\iclrfinalcopy

\usepackage{amsmath,amsfonts,amssymb}
\usepackage{graphicx}
\usepackage{booktabs}
\usepackage{hyperref}
\usepackage{url}
\usepackage{xcolor}
\usepackage{subcaption}
\usepackage{multirow}
\usepackage{tikz}
\usetikzlibrary{positioning,arrows.meta,fit,backgrounds,calc}

\title{Universal Transformers Need Memory:\\Depth-State Trade-offs in Adaptive Recursive Reasoning}

\author{Grigory Sapunov \\
Intento \\
\texttt{gs@inten.to}
}

\begin{document}
\maketitle

\begin{abstract}
We study learned memory tokens as a computational scratchpad for a single-block Universal Transformer with Adaptive Computation Time (ACT) on Sudoku-Extreme, a combinatorial reasoning benchmark. Memory tokens are empirically necessary: no configuration without them reaches non-trivial performance. The optimal count has a sharp lower threshold ($T{=}0$ always fails, $T{=}8$ reliably succeeds) followed by a stable plateau ($T{=}8$--$32$, $57.4\% \pm 0.7\%$ exact-match) and a dilution boundary at $T{=}64$. Under halt-side pressure ($\lambda$ warmup), mean halt drops monotonically with memory size across the plateau (from 11.6 at $T{=}8$ to 8.3 at $T{=}64$), showing that memory tokens and ponder depth substitute as resources at fixed accuracy.

We also identify a \textbf{router initialization trap} that causes the majority of training runs to fail: both default zero-bias and Graves' recommended positive bias settle into a shallow halt equilibrium the model cannot escape. Inverting the bias to $-3$ (``deep start'') eliminates the failure mode, and ablation shows the trap is inherent to ACT initialization rather than an artifact of our architecture.

With reliable training, ACT yields an order of magnitude lower seed variance than fixed-depth processing ($\pm 0.7$ vs $\pm 9.3$ pp); $\lambda$ warmup recovers 34\% of compute at matched accuracy; and attention heads specialize into memory readers, constraint propagators, and integrators across recursive depth. Code: \url{https://github.com/che-shr-cat/utm-jax}.
\end{abstract}

\section{Introduction}

Universal Transformers \citep{dehghani2019universal} apply a single transformer block iteratively, with Adaptive Computation Time \citep{graves2016adaptive} determining per-token processing depth. While theoretically appealing---arbitrary-depth reasoning with finite parameters---practical implementations have shown mixed results \citep{csordas2021devil}.

We investigate the role of learned memory tokens \citep{burtsev2020memory} in enabling recursive reasoning within this architecture, which we call \textbf{UTM} (Universal Transformer with Memory), using Sudoku-Extreme as a testbed. The title echoes \citet{darcet2024vision}, ``Vision Transformers Need Registers''---we present analogous evidence that our single-block UT needs memory tokens, based on extensive empirical failure without them.

Our contributions:
\begin{enumerate}
\item \textbf{Memory-token necessity, threshold, and depth-state substitution} (\S\ref{sec:memory}): In our single-block UT with ACT, no configuration without memory tokens achieves non-trivial performance across any tested seed, initialization, or depth mode. The optimal count shows a sharp threshold between $T{=}0$ (always fails) and $T{=}8$ (always succeeds) for 81-cell Sudoku, with a stable plateau through $T{=}32$. Memory tokens and ponder depth are substitutable resources at fixed accuracy: at $\lambda{=}0$, minimum halt steps decrease monotonically with $T$ ($17.7 \to 16.4 \to 15.5$); under halt-side pressure ($\lambda{=}0.001$+warmup), the trade-off becomes a clean monotonic curve, with mean halt dropping from $11.60$ at $T{=}8$ to $10.29$ at $T{=}32$ to $8.25$ at $T{=}64$ at near-constant accuracy on the plateau ($\sim 57\%$) and a 2pp drop at the dilution boundary. We note that other recursive architectures (TRM, HRM) solve similar tasks via different mechanisms---the necessity is architecture-specific.
\item \textbf{Router initialization trap} (\S\ref{sec:trap}): We identify that both default initialization (bias$=0$, $p \approx 0.5$) and Graves' recommended positive bias (bias$=1$, $p \approx 0.73$) create shallow-halt traps. In our setting, $>$70\% of runs fail to escape. We propose deep-start initialization (bias $= -3$, $p \approx 0.05$), which inverts the assumption and resolves the issue.
\item \textbf{ACT provides reliability and efficiency} (\S\ref{sec:act}): Fixed-depth processing with memory tokens achieves $53.4\% \pm 9.3\%$ EM (3 seeds)---high variance. ACT-enabled runs are more consistent ($56.9\% \pm 0.7\%$). Lambda warmup achieves $57.0\% \pm 1.1\%$ using 34\% fewer ponder steps---matching quality with significant compute savings.
\item \textbf{Diagnostic framework} (\S\ref{sec:trap}): Per-step router probability, step-weight distribution, and attention-mass logging that reveals head specialization and computation dynamics across recursive depth.
\end{enumerate}

\section{Architecture}

\subsection{Universal Transformer with Memory Tokens}

\begin{figure}[t]
\centering
\begin{tikzpicture}[
    block/.style={draw, rounded corners, minimum height=0.7cm, fill=blue!8, font=\small},
    tok/.style={draw, rounded corners, minimum width=0.9cm, minimum height=0.55cm, font=\scriptsize},
    arr/.style={-Stealth, thick},
    lab/.style={font=\scriptsize, text=gray!70},
    every node/.style={font=\small}
]
\node[tok, fill=purple!20] (m1) {$m_1$};
\node[tok, fill=purple!20, right=0.08cm of m1] (m2) {$m_2$};
\node[right=0.06cm of m2, font=\scriptsize] (mdots) {$\cdots$};
\node[tok, fill=purple!20, right=0.06cm of mdots] (mN) {$m_N$};
\node[tok, fill=orange!20, right=0.35cm of mN] (s1) {$s_1$};
\node[tok, fill=orange!20, right=0.08cm of s1] (s2) {$s_2$};
\node[right=0.06cm of s2, font=\scriptsize] (sdots) {$\cdots$};
\node[tok, fill=orange!20, right=0.06cm of sdots] (sL) {$s_L$};

\node[below=0.15cm of m2, font=\scriptsize, text=purple!60!black] {memory tokens};
\node[below=0.15cm of s2, font=\scriptsize, text=orange!60!black] {sequence tokens};

\coordinate (mem_left) at ($(m1.west)+(-0.1,0)$);
\coordinate (mem_right) at ($(mN.east)+(0.1,0)$);
\coordinate (seq_left) at ($(s1.west)+(-0.1,0)$);
\coordinate (seq_right) at ($(sL.east)+(0.1,0)$);

\node[block, above=0.7cm of $(m1)!0.5!(mN)$, fill=purple!10,
      minimum width={2.8cm}] (mememb) {\scriptsize Learned + Type Emb};
\node[block, above=0.7cm of $(s1)!0.5!(sL)$, fill=orange!10,
      minimum width={3.2cm}] (seqemb) {\scriptsize Token + Type Emb};

\draw[arr] (m1.north) -- (m1 |- mememb.south);
\draw[arr] (mN.north) -- (mN |- mememb.south);
\draw[arr] (s1.north) -- (s1 |- seqemb.south);
\draw[arr] (sL.north) -- (sL |- seqemb.south);

\coordinate (mid) at ($(mememb.north)!0.5!(seqemb.north)$);
\node[block, above=0.6cm of mid, minimum width=6.5cm, fill=blue!5] (input) {\scriptsize $[\text{mem}_1, \ldots, \text{mem}_N, \text{seq}_1, \ldots, \text{seq}_L]$ + Step Embedding};

\draw[arr] (mememb.north) -- (input.south -| mememb);
\draw[arr] (seqemb.north) -- (input.south -| seqemb);

\node[block, above=0.6cm of input, minimum width=6.5cm, fill=green!12, minimum height=0.9cm] (shared) {\textbf{Shared Block:} DerfNorm $\to$ MHA\,+\,RoPE $\to$ DerfNorm $\to$ SwiGLU};

\draw[arr] (input) -- (shared);

\node[block, right=0.5cm of shared, minimum width=1.2cm, fill=red!12, font=\scriptsize, align=center] (router) {ACT\\Router};
\draw[arr] (shared) -- (router);

\draw[arr, thick, blue!50] (shared.west) -- ++(-0.7,0) |- node[lab, left, pos=0.25] {$\times K$} (input.west);

\node[block, above=0.6cm of shared, minimum width=6.5cm, fill=yellow!12] (output) {ACT weighted blend $\sum w_k h_k$ ~~or~~ last step $h_K$};
\draw[arr] (shared) -- node[lab, right] {} (output);
\draw[arr, red!50, dashed] (router.north) |- node[lab, above, pos=0.7] {\scriptsize $p_k$, halt?} (output.east);

\node[block, above=0.5cm of output, minimum width=4cm, fill=gray!10] (proj) {Output Projection $\to$ predictions};
\draw[arr] (output) -- (proj);

\end{tikzpicture}
\caption{UTM architecture. Sequence tokens (orange) and memory tokens (purple) receive separate embeddings, are concatenated, and processed by a single weight-shared block iterated $K$ times. The ACT router outputs halting probability $p_k$ per token at each step; the final output is either the ACT-weighted blend or the last step's representation.}
\label{fig:arch}
\end{figure}
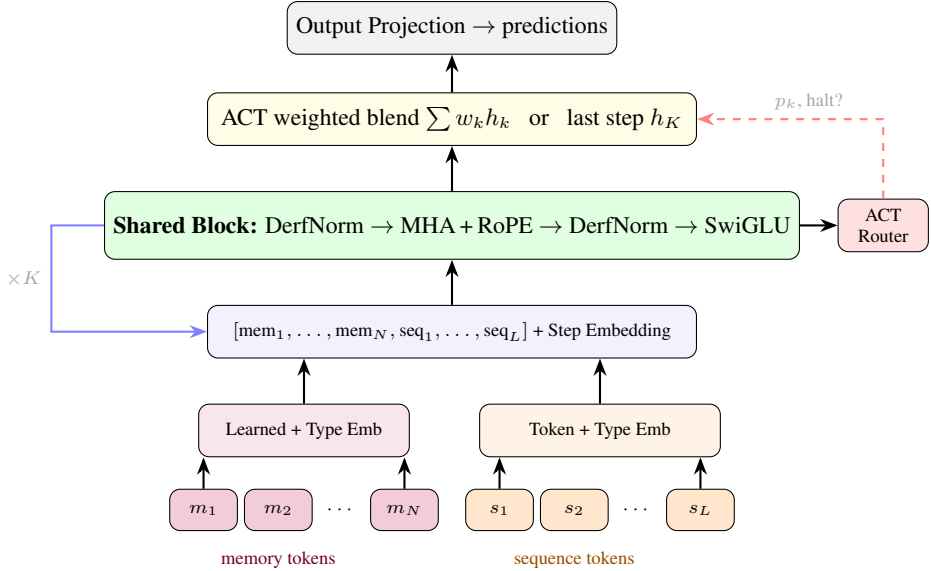

Our model applies a single \texttt{UniversalTransformerBlock} (pre-norm attention + SwiGLU FFN) iteratively for up to $K{=}18$ steps (Figure~\ref{fig:arch}). The full sequence at each step is $[\text{mem}_1, \ldots, \text{mem}_N, \text{seq}_1, \ldots, \text{seq}_L]$ with bidirectional attention.

\textbf{Components:} token + type embedding (memory vs sequence) + per-step learned positional embedding; multi-head attention with RoPE and QK-normalization; SwiGLU FFN with $8/3\times$ expansion; DerfNorm (our term for the normalization-free design using Derf \citep{chen2025derf}), where standard normalization layers are replaced by $\text{erf}(\alpha \cdot x + s)$ with learned per-feature $\alpha, s$; $N$ learned memory vectors with indexed RoPE positions (numbered registers). \textbf{Parameters:} 3.2M (hidden=512, heads=8, head\_dim=64, vocab=11).

\subsection{Adaptive Computation Time}

Following \citet{graves2016adaptive}, each token maintains cumulative halting probability. At step $k$, the router (linear + sigmoid) outputs $p_k \in (0,1)$. The output is a weighted blend: $\text{output} = \sum_{k=1}^{N} w_k \cdot h_k$, where $w_k = p_k$ for intermediate steps and $w_N = 1 - \sum_{k<N} p_k$ at halt. Ponder cost $\rho = N + R$ is minimized with coefficient $\lambda$.

When ACT is disabled, the model outputs only the final representation $h_K$ (standard weight-tied transformer).

\subsection{Deep-Start Initialization}

With default framework initialization (zero bias), the router computes $\sigma(W \cdot h + 0) \approx 0.5$, causing tokens to halt after ${\sim}2$ steps. \citet{graves2016adaptive} recommends initializing the halting bias to a \emph{positive} value ($b_h = 1$, giving $\sigma(\cdot) \approx 0.73$) to ``prevent very long sequences at the beginning of training''---which makes tokens halt even faster (${\sim}1$--$2$ steps). \citet{dehghani2019universal} does not specify halting initialization (Appendix C shows architecture but omits init details).

Both the default (bias$=0$, $p \approx 0.5$) and Graves' recommendation (bias$=1$, $p \approx 0.73$) produce shallow halting. We propose the opposite:

\textbf{Deep start:} bias $= -3$, giving $\sigma(W \cdot h - 3) \approx 0.05$. Tokens process all $K$ steps by default and learn to halt earlier. This inverts Graves' assumption: instead of preventing long sequences, we start with maximum depth and let the model discover where to stop. This is appropriate when the task requires significant depth---the cost of long initial sequences is small if the final learned policy uses them, whereas a shallow starting policy cannot easily discover that depth is needed.

\section{The Router Initialization Trap}
\label{sec:trap}

\subsection{Diagnosis}

We instrumented the ACT loop to log per-step router probability and router-specific gradient norm. Across 13 completed runs (planned 5 memory-token counts $\times$ 3 seeds; only $T{=}64$/$S{=}0$ was run for $T{=}64$ before we moved on; bias=0, $\lambda{=}0$, 4 epochs each):

\begin{table}[h]
\centering
\caption{Eval exact-match (\%) with standard initialization (bias=0). Bold = escaped the trap ($>$20\% EM). 4 of 13 completed runs succeed; seed 123 never escapes. Dashes mark configurations not run at this initialization.}
\label{tab:trap}
\begin{tabular}{lccccc}
\toprule
Seed & $T{=}0$ & $T{=}8$ & $T{=}16$ & $T{=}32$ & $T{=}64$ \\
\midrule
0   & 3.3  & 7.3  & \textbf{50.0} & 3.6  & \textbf{40.5} \\
42  & 2.7  & 3.3  & \textbf{50.6} & \textbf{57.2} & --- \\
123 & 4.7  & 4.1  & 3.6  & 4.4  & --- \\
\bottomrule
\end{tabular}
\end{table}

Diagnostic findings: (1) All runs start at $p \approx 0.48$--$0.52$, halt $= 2.0$. (2) By step 3k, all develop a shallow-halt pattern (halt $\approx 5$--$7$). (3) Escape correlates with a 10--45$\times$ spike in router gradient norm. (4) Stuck runs maintain router gradient $< 0.04$ for all 60k steps.

\begin{figure}[h]
\centering
\includegraphics[width=\textwidth]{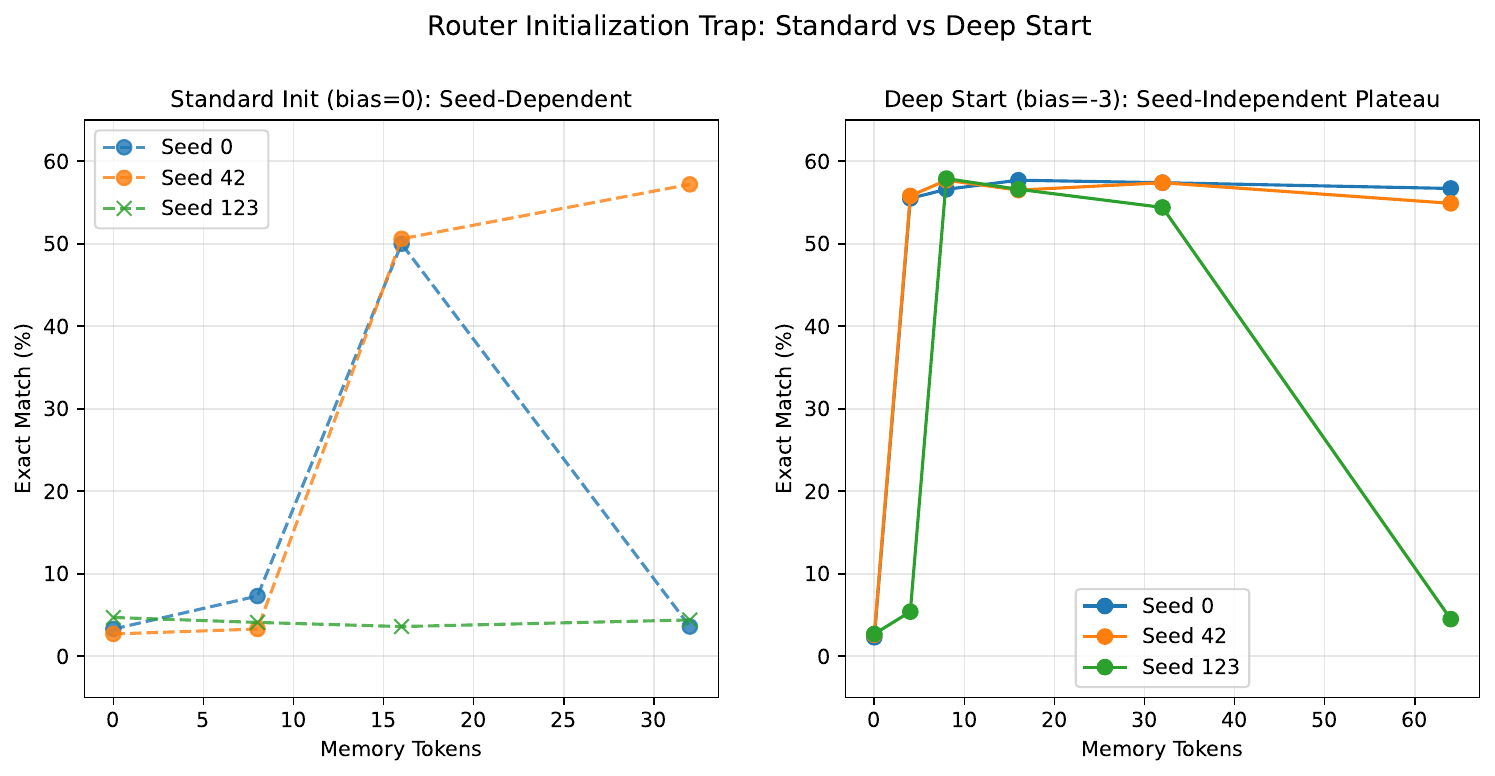}
\caption{Standard initialization (left) shows extreme seed sensitivity---same architecture, different outcomes by seed. Deep-start initialization (right) eliminates seed sensitivity: all seeds converge within the $T{=}8$--$32$ plateau.}
\label{fig:init_trap}
\end{figure}

\subsection{Deep Start Resolves the Trap}

With bias$=-3$, all previously-failing configurations at seed 123 succeed (Table~\ref{tab:fix}). Training dynamics change qualitatively: with bias=0, successful runs exhibit abrupt phase transitions; with bias$=-3$, accuracy rises smoothly.

\begin{table}[h]
\centering
\caption{Deep-start fix at seed 123 (the worst-performing seed from Table~\ref{tab:trap}).}
\label{tab:fix}
\begin{tabular}{lccl}
\toprule
$T$ & bias=0 & bias$=-3$ & Rescued? \\
\midrule
0  & 4.7\% & 2.7\% & No (memory needed) \\
4  & ---   & 5.4\% & No \\
8  & 4.1\% & \textbf{57.9\%} & Yes \\
16 & 3.6\% & \textbf{56.6\%} & Yes \\
32 & 4.4\% & \textbf{54.4\%} & Yes \\
64 & ---   & 4.5\% & No (attention dilution) \\
\bottomrule
\end{tabular}
\end{table}

\subsection{DerfNorm Is Not the Cause}

We verify the trap is not a DerfNorm artifact by replacing it with RMSNorm at the same seed and configuration ($T{=}16$, seed 42, bias=0): DerfNorm achieves 50.6\% EM (escapes); RMSNorm achieves \textbf{0.0\%} EM (halt stuck at 3.5, never escapes). The trap is \emph{worse} with standard normalization---it is inherent to ACT with standard initialization in our setting. This is consistent with \citet{chen2025derf}'s general finding that Derf outperforms RMSNorm across domains, though the ACT-escape dynamics we observe are specific to recursive architectures.

\section{Memory Tokens for Recursive Reasoning}
\label{sec:memory}

\subsection{Task and Setup}

\textbf{Sudoku-Extreme} \citep{sudokuextreme}: 9$\times$9 puzzles, extreme difficulty (17--24 givens). 3.83M train / 423K test. Encoder-style prediction of all 81 cells simultaneously. All results: hidden=512, heads=8, max\_ponder=18, batch=256, AdamW (lr=$3 \times 10^{-4}$, cosine decay), EMA (0.999), 4 epochs, bias$=-3$, $\lambda{=}0$ unless noted.

\subsection{Memory-Token Curve}

\begin{figure}[t]
\centering
\includegraphics[width=0.85\textwidth]{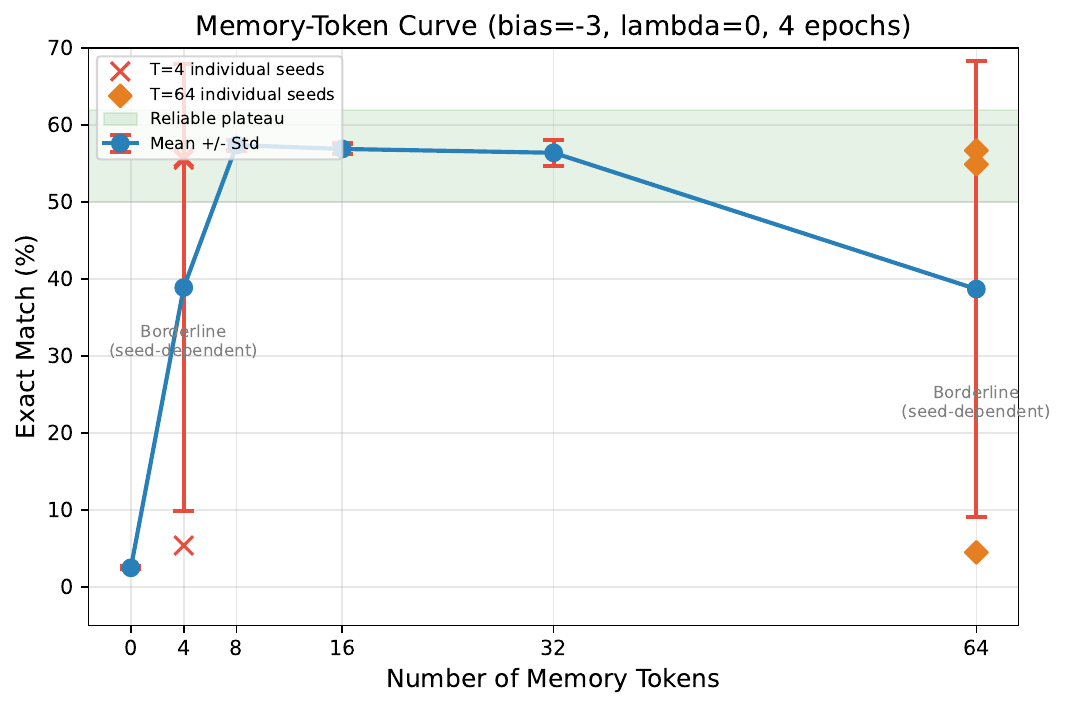}
\caption{Memory-token curve with deep-start initialization. $T{=}0$ always fails; $T{=}4$ and $T{=}64$ are borderline (seed-dependent); $T{=}8$--$32$ is a stable plateau ($57.4 \pm 0.7\%$).}
\label{fig:curve}
\end{figure}

\begin{table}[h]
\centering
\caption{Memory-token curve with deep-start initialization ($\lambda{=}0$, 4 epochs). All rows at 3 seeds (S=0, 42, 123). Mean halt is reported only for non-bimodal rows; halt range covers all 3 seeds.}
\label{tab:curve}
\begin{tabular}{lccccccc}
\toprule
$T$ & $S{=}0$ & $S{=}42$ & $S{=}123$ & EM Mean $\pm$ Std & Halt Mean & Halt Range \\
\midrule
0   & 2.3\% & 2.6\%  & 2.7\%  & $2.5 \pm 0.2$\% & 7.83 & 6.9--8.5 \\
4   & 55.5\% & 55.8\%    & 5.4\%  & --- (bimodal)  & --- & 7.0--17.7 \\
\textbf{8}  & \textbf{56.6\%} & \textbf{57.7\%} & \textbf{57.9\%} & $\mathbf{57.4 \pm 0.7}$\% & \textbf{17.87} & 17.7--18.0 \\
\textbf{16} & \textbf{57.7\%} & \textbf{56.5\%} & \textbf{56.6\%} & $\mathbf{56.9 \pm 0.7}$\% & \textbf{17.07} & 16.4--18.0 \\
\textbf{32} & \textbf{57.4\%} & \textbf{57.4\%} & \textbf{54.4\%} & $\mathbf{56.4 \pm 1.7}$\% & \textbf{17.03} & 15.5--18.0 \\
64  & 56.7\% & 54.9\%    & 4.5\%  & --- (bimodal)  & --- & 6.4--16.1 \\
\bottomrule
\end{tabular}
\end{table}

Key findings: (1) \textbf{Sharp threshold between $T{=}0$ and $T{=}8$}: $T{=}0$ always fails ($2.5\%$, 3 seeds). $T{=}4$ is borderline---succeeds at 2 of 3 seeds (55.5--55.8\%) but fails at the third (5.4\%), exhibiting the same seed sensitivity as $T{=}64$. $T{=}8$ always succeeds ($57.4 \pm 0.7\%$, 3 seeds). The minimum reliable scratchpad for 81-cell Sudoku is 8 tokens (${\sim}1$ per 10 cells). (2) \textbf{Stable plateau}: $T{=}8$: $57.4{\pm}0.7$\%, $T{=}16$: $56.9{\pm}0.7$\%, $T{=}32$: $56.4{\pm}1.7$\%. (3) \textbf{$T{=}0$ fails with deep start}: memory tokens are necessary in this architecture. We note that TRM \citep{jolicoeur2025trm} and HRM \citep{wang2025hrm} solve Sudoku without memory tokens via autoregressive answer improvement---the necessity is architecture-specific to our single-block UT.

\subsection{Memory-Depth Tradeoff}
\label{sec:tradeoff}

Within the plateau, memory tokens and ponder depth function as substitutable resources: as $T$ increases, the model can solve the task at the same accuracy with fewer ponder steps. The clearest evidence is the per-seed minimum halt step in Table~\ref{tab:curve}: $T{=}8$ saturates the 18-step ceiling at all 3 seeds (halt $\in [17.7, 18.0]$), $T{=}16$ admits some headroom ($[16.4, 18.0]$), and $T{=}32$ admits more ($[15.5, 18.0]$). The substitution is real but seed-dependent --- at $T{=}32$, $S{=}0$ exploits the headroom (halt $= 15.5$) while $S{=}42$ stays saturated (halt $= 18.0$), and different seeds find different equilibria along the same accuracy plateau.

\textbf{Mean halt is dominated by the depth ceiling at $\lambda{=}0$.} Across the plateau, mean halt is $17.87$, $17.07$, $17.03$ for $T = 8, 16, 32$ respectively (Table~\ref{tab:curve}). The compression is small in the mean because $\lambda{=}0$ in this sweep --- with no penalty for using deep ponder, the model saturates the ceiling whenever it can, and the substitution surfaces only in the per-seed minima. The trade-off is more visible under halt-side pressure: with $\lambda{=}0.001$ and a 20k-step warmup at $T{=}16$ across 3 seeds (\S\ref{sec:act}), mean halt drops from $17.07$ to $\sim 11.4$ at unchanged accuracy ($57.0 \pm 1.1\%$ vs $56.9 \pm 0.7\%$). This is the canonical depth-state trade-off form: fixed accuracy, reduced compute, achieved by giving the router something to push against.

\begin{figure}[t]
\centering
\includegraphics[width=\textwidth]{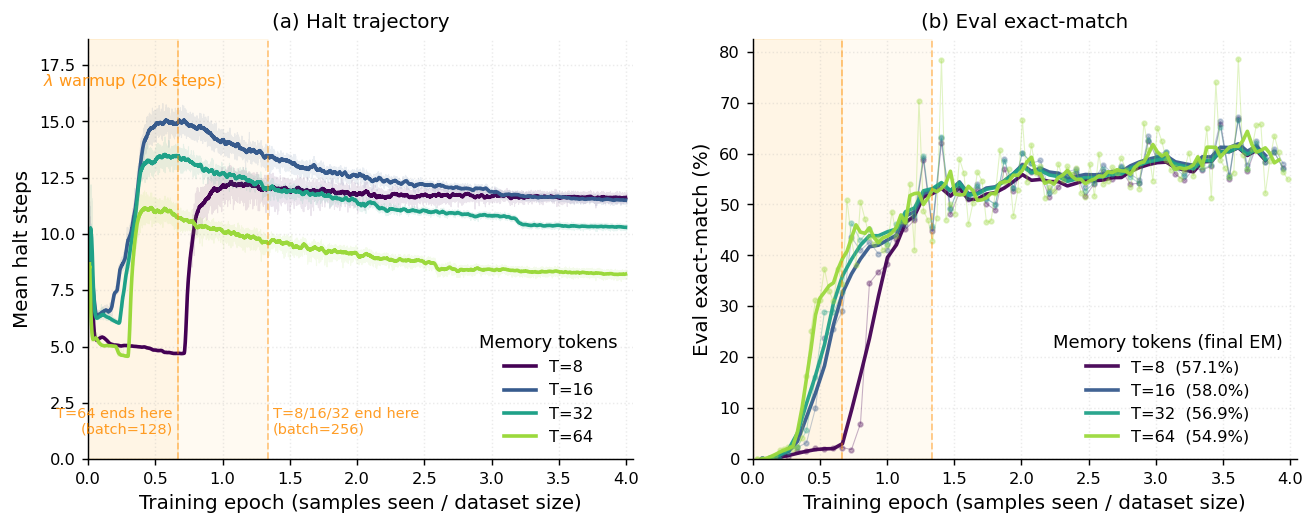}
\caption{Training trajectories for the four runs of Table~\ref{tab:tradeoff} ($\lambda{=}0.001$ + 20k-step warmup, $S{=}0$). \textbf{(a)} Mean halt steps. \textbf{(b)} Held-out (eval) exact-match. The x-axis is samples seen ($=$ step $\times$ batch size) so the four runs are directly comparable across the same 4-epoch budget. The 20k-step $\lambda$-warmup interval translates to different positions on this axis depending on batch size: $T{=}8/16/32$ used batch~256 and finish warmup at ${\sim}1.34$ epochs (right dashed line); $T{=}64$ used batch~128 (HBM constraint) and finishes warmup at ${\sim}0.67$ epochs (left dashed line). The shaded region marks the union of both warmup intervals. \textbf{Halt panel:} after warmup, mean halt monotonically settles to lower values for larger $T$ (final values 11.6, 11.5, 10.3, 8.2), supporting the depth--state substitution claim. Two subtleties are visible at the start of training. (i) Deep-start initialization (bias$=-3$) makes all four runs begin at the halt ceiling (18), but within the first ${\sim}10$ training steps the router collapses to a shallow equilibrium near halt~$=$~5--7 as soon as gradients flow. (ii) Larger-$T$ runs escape this shallow regime during warmup (climbing to ${\sim}10$--$15$ as compute pressure builds), while $T{=}8$ stays in the shallow regime for nearly a full epoch before climbing --- the smallest memory has the least to gain from extra ponder, so the cost--benefit balance shifts toward pondering more only when $\lambda$ becomes non-trivial. \textbf{Eval panel:} faint markers are raw evals (per-1000-step on a held-out sample), thick lines are a 5-eval rolling mean. All four runs converge to a similar plateau ($\approx 55$--$60\%$ EM, with eval-to-eval variance of several pp); the trajectory shows that the substitution is essentially free in quality across $T \in \{8, 16, 32, 64\}$. The 2pp dilution gap reported in Table~\ref{tab:tradeoff} is the difference between final eval points and is at the level of single-eval noise here, so it is not visually separable on the trajectory --- which is itself a useful sanity check on the substitution claim.}
\label{fig:halt_trajectory}
\end{figure}

\textbf{The full curve under halt-side pressure.} A direct $T \times \lambda$+warmup sweep at $S{=}0$ confirms the trade-off in its canonical form (Table~\ref{tab:tradeoff}, Figure~\ref{fig:halt_trajectory}): mean halt decreases monotonically from $11.60$ at $T{=}8$ to $11.50$ at $T{=}16$ to $10.29$ at $T{=}32$ to $8.25$ at $T{=}64$, with EM essentially constant across the plateau ($57.09 \to 58.00 \to 56.87$) and a 2pp drop at the dilution boundary ($T{=}64$, $54.91\%$). Compared to $\lambda{=}0$ (Table~\ref{tab:curve}, $S{=}0$ column), halt is compressed by 33--40\% across the plateau at unchanged or slightly improved accuracy.

\textbf{Total compute is approximately conserved, not strictly fungible.} Total token-step compute (memory plus sequence tokens times mean halt) grows mildly with $T$ across the plateau ($1032 \to 1116 \to 1163$ token-steps for $T = 8, 16, 32$, ${\sim}13\%$ growth), reaching $1196$ at $T{=}64$. Memory and depth substitute in halt count and per-step bandwidth, but total operations are not strictly preserved.

\textbf{The dilution boundary is compatible with halt-side pressure.} At $T{=}64$, halt drops further (8.25) and accuracy is 2pp below the plateau ($54.91\%$ vs ${\sim}57\%$). With only one seed in this sweep, we cannot rule out single-seed noise from the 2pp gap alone (the eval trajectory in Figure~\ref{fig:halt_trajectory}b shows the four runs intermingled across the plateau). However, the magnitude and direction are consistent with the bimodal collapse observed at $T{=}64$ under $\lambda{=}0$ across multiple seeds (Table~\ref{tab:curve}); read together, they support treating $T{=}64$ as a dilution boundary --- an architectural property at large $T$ --- rather than an artifact of saturated halting at $\lambda{=}0$.

\begin{table}[h]
\centering
\caption{Depth--state trade-off curve under halt-side pressure ($\lambda{=}0.001$ + 20k-step warmup, deep-start, $S{=}0$, 4 epochs). Halt decreases monotonically across the plateau and into the dilution zone; accuracy is essentially constant across $T{=}8$--$32$ and falls 2pp at $T{=}64$. Total token-steps $= (T + 81) \times \text{Mean Halt}$.}
\label{tab:tradeoff}
\begin{tabular}{lccc}
\toprule
$T$ & Mean Halt & EM & Total token-steps \\
\midrule
8  & 11.60 & 57.09\% & 1032 \\
16 & 11.50 & 58.00\% & 1116 \\
32 & 10.29 & 56.87\% & 1163 \\
64 & 8.25  & 54.91\% & 1196 \\
\bottomrule
\end{tabular}
\end{table}

\subsection{Fixed-Depth vs ACT Processing}

\begin{table}[h]
\centering
\caption{Fixed-depth vs ACT ablation ($T{=}16$, bias$=-3$, 3 seeds). ACT's weighted blend provides consistency; fixed-depth is sensitive to seed.}
\label{tab:fixed}
\begin{tabular}{lcccc}
\toprule
Config & $S{=}0$ & $S{=}42$ & $S{=}123$ & Mean $\pm$ Std \\
\midrule
ACT enabled, $\lambda{=}0$ & 57.7\% & 56.5\% & 56.6\% & $56.9 \pm 0.7$\% \\
ACT, $\lambda{=}0.001$+warmup & 58.0\% & 57.0\% & 55.9\% & $57.0 \pm 1.1$\% \\
ACT disabled (fixed-18) & 52.0\% & 44.9\% & 63.4\% & $53.4 \pm 9.3$\% \\
\bottomrule
\end{tabular}
\end{table}

Fixed-depth processing achieves comparable mean EM ($53.4\%$) but with \textbf{much higher variance}: EM ranges from 44.9\% to 63.4\% across 3 seeds ($\pm 9.3\%$), compared to 56.5--57.7\% for ACT ($\pm 0.7\%$)---an order-of-magnitude difference in seed sensitivity. We attribute this to the output mechanism: ACT blends representations across all $K$ steps ($\sum w_k h_k$), averaging out seed-dependent variation in individual steps; fixed-depth relies entirely on the final step's representation $h_K$, which may be more sensitive to initialization-dependent optimization trajectories since it is a single snapshot rather than an average over steps. Lambda warmup combines ACT's reliability ($57.0 \pm 1.1\%$) with 34\% compute savings.

\subsection{How Memory Tokens Function: Attention Analysis}

\begin{figure}[t]
\centering
\includegraphics[width=\textwidth]{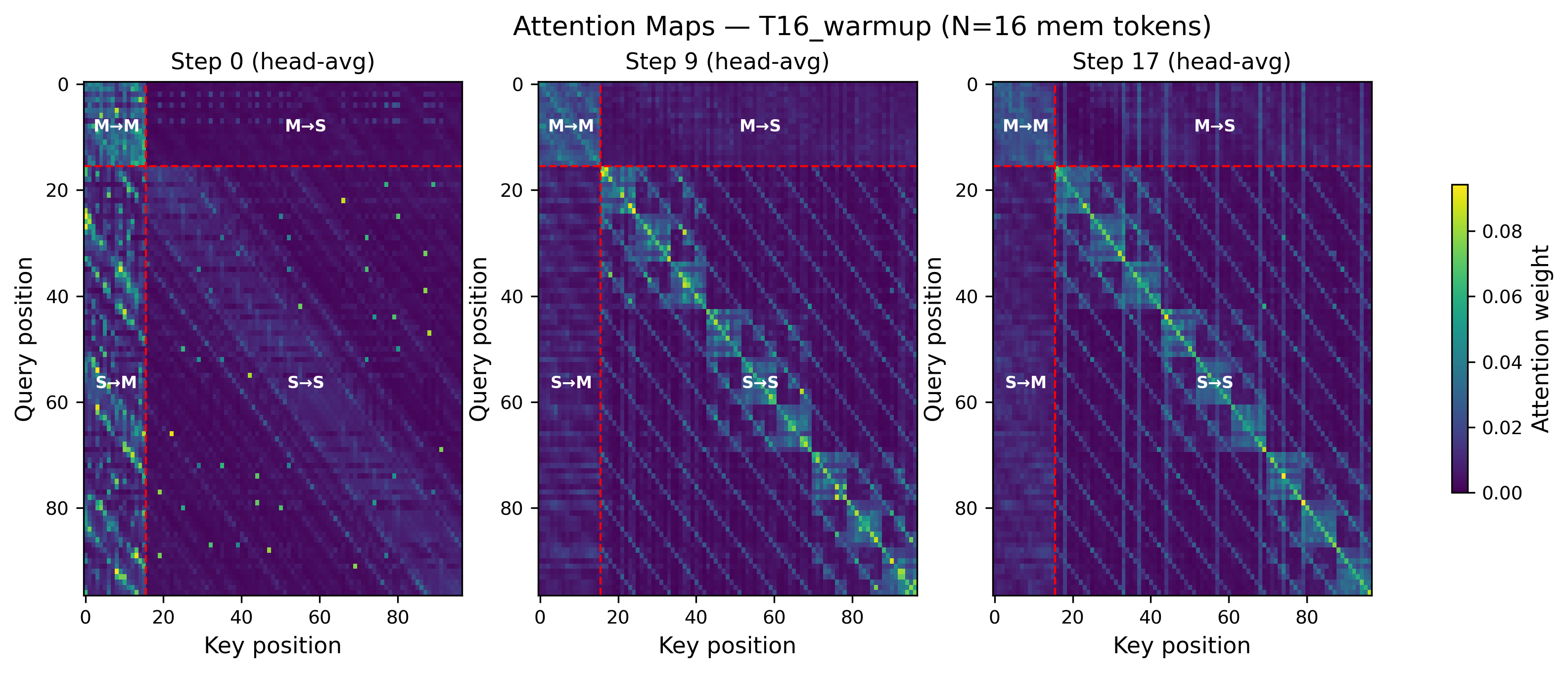}
\caption{Attention maps at steps 0, 9, and 17 (head-averaged, $T{=}16$). Red lines delineate memory/sequence quadrants. S$\to$S attention develops block-diagonal structure matching Sudoku constraints. S$\to$M shows sequence tokens querying specific memory slots.}
\label{fig:attn_maps}
\end{figure}

\begin{figure}[t]
\centering
\includegraphics[width=\textwidth]{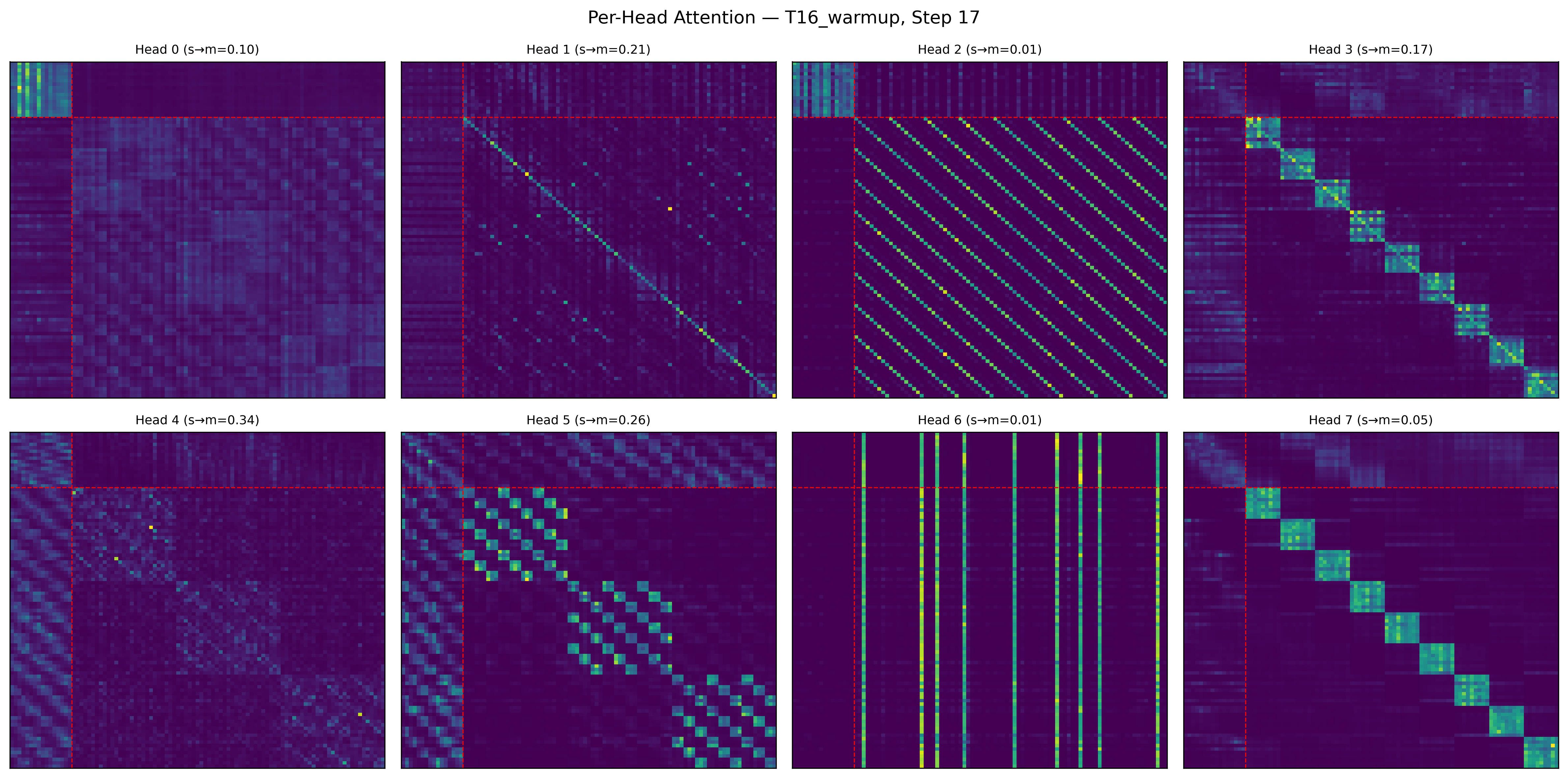}
\caption{Per-head attention at step 17 ($T{=}16$). Each head's s$\to$m fraction is shown. H4 (0.34) and H1 (0.24) are memory-focused; H2 and H6 (0.01) are pure puzzle-constraint heads with periodic S$\to$S structure; H5 (0.21) mixes column-attention stripes with memory reading. See text for per-head analysis.}
\label{fig:attn_heads}
\end{figure}

Attention evolves across depth: at step 0, attention is diffuse; by step 9, S$\to$S develops block-diagonal patterns matching Sudoku row/column/box constraints; by step 17, the structure is highly refined (Figure~\ref{fig:attn_maps}).

Heads specialize into distinct roles (Figure~\ref{fig:attn_heads}, Table~\ref{tab:heads}). The full attention quadrant breakdown reveals asymmetric memory usage:

\begin{table}[h]
\centering
\caption{Per-head attention quadrants at step 17 ($T{=}16$, best model). Each row sums to ${\sim}1$ for sequence queries (s$\to$m + s$\to$s) and memory queries (m$\to$m + m$\to$s) separately.}
\label{tab:heads}
\scriptsize
\begin{tabular}{lccccll}
\toprule
Head & s$\to$m & s$\to$s & m$\to$m & m$\to$s & Visual pattern & Role \\
\midrule
H0 & 0.10 & 0.90 & \textbf{0.73} & 0.27 & Diffuse S$\to$S & Memory self-org \\
H1 & 0.21 & 0.79 & 0.19 & \textbf{0.81} & S$\to$M bands & Memory writer \\
H2 & 0.01 & \textbf{0.99} & \textbf{0.74} & 0.26 & Diagonal S$\to$S bands & Constraint prop. \\
H3 & 0.18 & 0.82 & 0.20 & \textbf{0.80} & Structured S$\to$S blocks & Memory writer \\
H4 & \textbf{0.34} & 0.66 & 0.45 & 0.55 & Strong S$\to$M & Memory reader \\
H5 & 0.22 & 0.78 & 0.24 & 0.77 & Checkerboard S$\to$S & Mixed read + constraint \\
H6 & 0.01 & \textbf{1.00} & 0.00 & \textbf{1.00} & Vertical stripes S$\to$S & Column constraint \\
H7 & 0.06 & 0.94 & 0.26 & 0.74 & Block-diagonal S$\to$S & Localized constraint \\
\bottomrule
\end{tabular}
\end{table}

Three functional groups emerge: \textbf{Memory readers} (H4, H5: high s$\to$m, sequence queries memory), \textbf{memory writers} (H1, H3: high m$\to$s, memory broadcasts to sequence), and \textbf{constraint propagators} (H2, H6: s$\to$m $\approx 0$, structured S$\to$S patterns). H0 and H2 show high m$\to$m (memory self-attention, ${\sim}0.73$), suggesting internal memory coordination. H6 is striking: both s$\to$s and m$\to$s are ${\sim}1.0$---it propagates puzzle constraints while memory tokens passively observe the sequence.

In the trapped model (0\% EM), no head develops s$\to$m $> 0.16$, attention is disorganized, and the constraint patterns seen in H2/H6 are absent.

\subsection{Small-Dataset Generalization}

Following the TRM protocol (1000 training puzzles, 1000 augmentations, full 423K test set):

\begin{table}[h]
\centering
\caption{Small-dataset generalization. The model memorizes but does not generalize.}
\begin{tabular}{lcc}
\toprule
Config & Train EM & Eval EM (unseen) \\
\midrule
$\lambda{=}0$, wd=0.01 & 100\% & 8.1\% \\
$\lambda{=}0.001$+warmup, wd=0.1 & 99.6\% & 6.0\% \\
\midrule
TRM (7M params) & --- & 87.4\% \\
\bottomrule
\end{tabular}
\end{table}

The model memorizes perfectly but achieves only 6--8\% on unseen puzzles, even with TRM-matched regularization. This suggests that TRM's autoregressive answer-improvement mechanism may be better suited for algorithmic generalization, though the architectures differ in multiple ways and isolating the causal factor remains future work.

\section{Making ACT Efficient}
\label{sec:act}

\subsection{Lambda Warmup}

\begin{table}[h]
\centering
\caption{Lambda comparison ($T{=}16$, bias$=-3$). $\lambda{=}0$ baseline and $\lambda{=}0.001$+warmup measured at 3 seeds; the $\lambda{=}0.001$/no-warmup row is a single seed (S=123, illustrative collapse). Warmup achieves 34\% compute savings.}
\label{tab:lambda}
\begin{tabular}{lccccc}
\toprule
$\lambda$ & Warmup & Mean EM $\pm$ Std & Mean Halt & Savings \\
\midrule
0     & ---      & $56.9 \pm 0.7$\% & 16.4--18.0 & baseline \\
0.001 & none     & 3.8\% (S=123) & 3.7  & collapsed \\
\textbf{0.001} & \textbf{20k steps} & $\mathbf{57.0 \pm 1.1}$\textbf{\%} & \textbf{11.4} & \textbf{$-$34\%} \\
\bottomrule
\end{tabular}
\end{table}

Direct application of the ponder penalty collapses halting even with deep start. Lambda warmup resolves this: the model establishes deep processing during the warmup phase, then the penalty compresses computation to halt $\approx 11$. Across 3 seeds, warmup achieves $57.0 \pm 1.1\%$ EM---matching $\lambda{=}0$ quality ($56.9 \pm 0.7\%$) with 34\% fewer steps.

\subsection{Inference Beyond Trained Depth}

\begin{figure}[h]
\centering
\includegraphics[width=\textwidth]{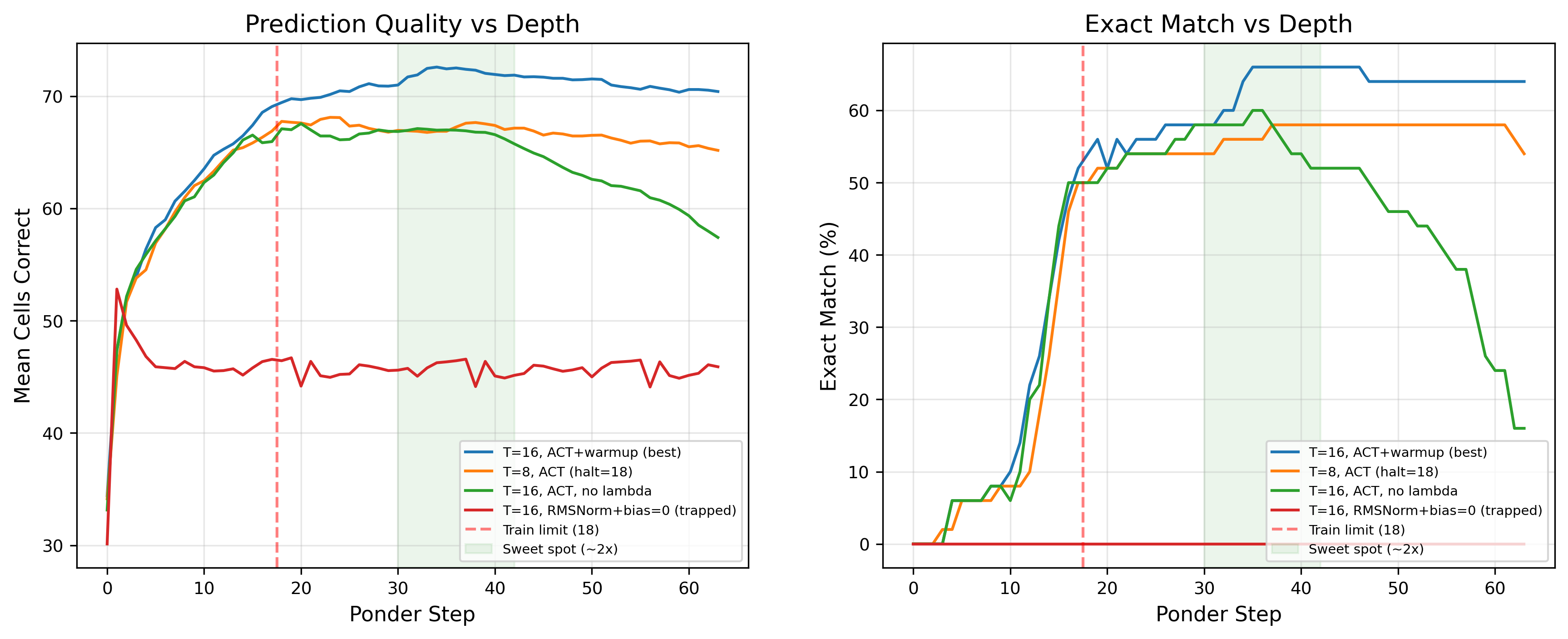}
\caption{Extended inference (64 steps, trained on 18). The lambda-warmup model peaks at step 36 (66\% EM, +14pp over step 17) then gradually degrades but never crashes (64\% at step 63). Green band marks the ${\sim}2\times$ sweet spot. Other models plateau by step 28. The trapped model is flat.}
\label{fig:extended}
\end{figure}

Models generalize to more ponder steps than trained (step embeddings wrap modularly). Running the lambda-warmup model for 36 steps (2$\times$ training depth) yields 66\% EM, up from 52\% at step 17---a 14pp gain from inference-time compute alone, with no retraining. Beyond 36 steps, quality plateaus and slowly degrades (64\% at step 48, 64\% at step 63), but never crashes---the model degrades gracefully. The sweet spot is ${\sim}2\times$ training depth.

Other models improve modestly (+4--6pp) and plateau by step 28. The practical implication: one can train with aggressive lambda for efficiency (halting at ${\sim}11$ steps), then run inference at $2\times$ depth when accuracy matters, recovering quality that the penalty compressed away.

\section{Related Work and Architectural Convergence}

The architecture we study sits at the intersection of three threads of research that developed largely independently and have re-engaged in 2025--2026 around a similar set of commitments: applying a single weight-shared block recursively, with adaptive halting, and some form of persistent state across recursion. We trace each thread, then map the design space they collectively occupy.

\subsection{Universal Transformers and Adaptive Computation Time}

Adaptive Computation Time was introduced by \citet{graves2016adaptive} for recurrent neural networks as a mechanism for the network to learn how much computation to spend on each input. Each step produces a halting probability via a sigmoid output; cumulative halting probabilities trigger early termination, and a remainder term provides the gradient path that lets the network learn when to stop.

\citet{dehghani2019universal} combined ACT with the Transformer architecture in the Universal Transformer (UT), applying a single weight-shared block iteratively up to a maximum depth. UTs decouple parameter count from depth and provide a path to architectures with theoretical universality properties. Empirically, however, ACT in UTs has shown mixed results---for example, \citet{csordas2021devil} implement Universal Transformers as ``simply Transformers with shared weights between layers, without adaptive computation time'' (\S2.2), reflecting a broader trend of dropping ACT from UT variants in practice. Our findings on the router initialization trap (\S\ref{sec:trap}) suggest that some of these difficulties may reflect optimization pathologies in standard initialization rather than fundamental limitations of the architecture.

PonderNet \citep{banino2021pondernet} reformulates ACT's halting policy as a probabilistic model with a geometric distribution over halting steps, motivated by the observation that ``ACT is notably unstable and sensitive to the choice of a hyper-parameter that trades-off accuracy and computation cost. Additionally, the gradient for the cost of computation can only back-propagate through the last computational step, leading to a biased estimation of the gradient.'' PonderNet's reformulation provides unbiased gradient estimates while remaining fully differentiable.

\citet{saunshi2025reasoning} provide theoretical and empirical evidence that ``for many synthetic reasoning problems like addition, $p$-hop induction, and math problems, a $k$-layer transformer looped $L$ times nearly matches the performance of a $kL$-layer non-looped model, and is significantly better than a $k$-layer model.'' This supports the case for weight-shared depth recurrence as a general principle, of which our minimal architecture is one instantiation.

\subsection{Memory Tokens and Registers}

Memory tokens were introduced by \citet{burtsev2020memory} in the Memory Transformer, where a small set of learnable vectors are prepended to the input sequence to ``store non-local representations'' and create a ``memory bottleneck for the global information''---a learned global scratchpad attended to by all sequence positions through standard self-attention. \citet{bulatov2022recurrent} extended the mechanism for long-context recurrence in the Recurrent Memory Transformer (RMT), where memory tokens are passed across sequence segments to enable processing beyond the model's nominal context window.

\citet{darcet2024vision} independently rediscovered the same primitive for vision transformers under the name \emph{registers}, motivated by the observation that ``high-norm tokens appearing during inference primarily in low-informative background areas of images'' are repurposed by the model for internal computations. Their fix was to append additional tokens to the token sequence---independent of the input image---that the model can learn to use as registers, which absorbs the high-norm pathology and restores clean attention maps. The two genealogies---memory tokens for long-context language modeling, registers for vision transformer interpretability---describe the same architectural mechanism: prepended learnable tokens that participate fully in attention but are not tied to any input position.

Our work applies this mechanism inside the depth-recurrent loop of a UT. Memory tokens in our architecture persist across ponder steps within a single forward pass and serve as recursion-updated working memory rather than long-context summary or interpretability scaffolding.

\subsection{Recursive Reasoning on Combinatorial Benchmarks}

A recent line of work has focused on weight-shared recursive computation under the name \emph{recursive reasoning}, motivated by performance on combinatorial puzzle benchmarks (Sudoku, mazes, ARC-AGI).

HRM \citep{wang2025hrm} introduced two interdependent recurrent modules---a high-level module for ``slow, abstract planning'' and a low-level module for ``rapid, detailed computations''---both implemented as encoder-only Transformer blocks with identical architectures (4 layers each, per \citealp{jolicoeur2025trm}), with adaptive halting via Q-learning and deep supervision across up to 16 segments. HRM achieves strong small-data generalization with ${\sim}1$K training examples plus heavy augmentation. Its Related Work explicitly positions the architecture within the Universal Transformer / ACT lineage, noting that ``Like earlier neural reasoning algorithms including the Universal Transformer, HRM is computationally universal when given sufficient memory and time constraints'' (\S6).

TRM \citep{jolicoeur2025trm} simplifies HRM's two-network design to ``a single network with only 2 layers.'' The architecture maintains two recursive latent variables: a proposed solution $y$ and a latent reasoning feature $z$ (the paper explicitly renames HRM's $z_H$ and $z_L$ for the TRM algorithm: ``there is simply an input $x$, a proposed solution $y$ (previously called $z_H$), and a latent reasoning feature $z$ (previously called $z_L$)''). The MLP-Mixer variant (TRM-MLP, 5M parameters) reaches 87.4\% on Sudoku-Extreme; the attention variant (TRM-Att, 7M parameters) reaches 74.7\% on Sudoku-Extreme and 85.3\% on Maze-Hard. TRM trains via deep supervision combined with full back-propagation through the recursion (rather than HRM's 1-step gradient approximation) and uses a simplified halting mechanism that drops the continue Q-value entirely.

URM \citep{urm2025} explicitly identifies its architecture as a Universal Transformer variant. Its \S2.2 (``Universal Transformer'') opens by attributing the architecture to \citet{dehghani2019universal}: ``The Universal Transformer (UT) extends the standard Transformer by introducing recurrent computation over depth. Instead of stacking $L$ distinct layers, the UT applies a single transition block repeatedly to refine token representations.'' URM applies UT+ACT in an encoder-style (non-causal) design with a ConvSwiGLU MLP and Truncated Backpropagation Through Loops (TBPTL), reaching 77.6\% on Sudoku and 53.8\% pass@1 on ARC-AGI-1.\footnote{The URM paper describes the architecture as ``decoder-only,'' but the public implementation uses bidirectional self-attention (\texttt{causal=False} in \texttt{models/urm/urm.py}); we describe it accordingly. This matches the HRM and TRM lineage, which likewise refine a fixed-length puzzle sequence rather than generate autoregressively.} URM positions its contribution as identifying which mechanisms within the UT family drive performance, finding that ``recurrent inductive bias'' and the ConvSwiGLU nonlinearity matter more than elaborate architectural designs.

Across these three works, distinct architectural commitments are added to the underlying UT-with-ACT skeleton: HRM contributes hierarchical state, deep supervision, and the 1-step gradient approximation; TRM contributes the single-network simplification, full-recursion backpropagation, and the renaming-and-clarification of HRM's two latents into a clean ($x$, $y$, $z$) structure; URM contributes ConvSwiGLU, TBPTL, and explicit positioning within the UT lineage. Our work occupies the simpler corner of this space---single block, no per-position summed latent state, single-pass training---and isolates what this minimal configuration requires to function.

\subsection{Recursive Transformers at Language-Modeling Scale}

A complementary line of recent work has scaled weight-shared recursive computation to language-modeling pretraining, providing convergent evidence about the role of memory mechanisms.

\citet{geiping2025recurrent} (Huginn) study ``a novel language model architecture that is capable of scaling test-time computation by implicitly reasoning in latent space.'' Their model ``iterat[es] a recurrent block, thereby unrolling to arbitrary depth at test-time,'' and is scaled to 3.5 billion parameters and 800 billion tokens. They show the model can improve reasoning performance with additional inference-time compute ``up to a computation load equivalent to 50 billion parameters.'' This is the canonical LM-scale recurrent-depth pretraining work and demonstrates that the depth-as-third-axis premise of UT/ACT survives into the modern LM regime.

\citet{zhu2025ouro} (Ouro) extend this thread with ``a family of pre-trained Looped Language Models (LoopLM) that \ldots build reasoning into the pre-training phase through (i) iterative computation in latent space, (ii) an entropy-regularized objective for learned depth allocation, and (iii) scaling to 7.7T tokens.'' Ouro's halting mechanism is explicitly linked to PonderNet: the authors describe an exit gate that runs in parallel with the LM head at each step and aligns with adaptive-computation methods like PonderNet, which also optimize an ELBO objective for dynamic halting. At 1.4B and 2.6B parameters, Ouro models match the results of up to 12B SOTA LLMs across a wide range of benchmarks. Importantly, Ouro reports that ``recurrence does not increase raw knowledge storage (approximately 2 bits per parameter for both looped and non-looped models) but dramatically enhances knowledge manipulation capabilities on tasks requiring fact composition and multi-hop reasoning.'' Ouro carries information through the hidden state alone, without any auxiliary memory mechanism.

\citet{bae2025mor} propose Mixture-of-Recursions (MoR), ``a unified framework that combines [parameter sharing and adaptive computation] inside a single Recursive Transformer. MoR reuses a shared stack of layers across recursion steps to achieve parameter efficiency, while lightweight routers enable adaptive token-level thinking.'' MoR offers an alternative path to per-token compute allocation that is orthogonal to memory-token augmentation.

\citet{yu2026mesh} (MeSH) diagnose the same fundamental issue we motivate---that the single hidden state in a recursive transformer is overloaded, forced to simultaneously carry persistent context and transient computation. They quantify this with three observables: a skewed computational pattern (the first loop performs most of the work, subsequent loops contribute negligibly), representational stagnation (high CKA similarity between consecutive loop states), and loop representational collapse (rapid singular-value decay indicating low-dimensional collapse). Their solution, Memory-as-State-Highways, ``replaces the overloaded hidden state with an explicit memory buffer governed by lightweight, step-wise routers.'' MeSH-enhanced recursive transformers at the 1.4B Pythia scale, with 33\% fewer non-embedding parameters than the non-recursive counterpart, ``improv[e] average downstream accuracy by +1.06\%.'' Architecturally, MeSH slots are full-sequence-shaped tensors (closer to TRM's per-position latents than to our memory tokens), and the central architectural innovation is breaking weight-tying for the routers---each iteration uses different routing parameters---which is what enables the functional specialization across iterations they observe.

\citet{frey2026adaptive} combine adaptive per-layer looping with gated memory banks at the ${\sim}200$M parameter scale on language modeling. Their architecture pairs PonderNet-style halting with two types of static learnable memory: local memory banks per layer and a global memory bank shared across layers, both retrieved via key-value attention with input-dependent gating. They report a functional dissociation: looping primarily benefits mathematical reasoning (a Loop-3 model reduces math BPB from 2.163 to 1.687, a 22\% reduction), while memory banks help recover commonsense performance, and the combination outperforms an iso-FLOP baseline with three times the layers on math benchmarks. They frame this as a distinction between knowledge \emph{manipulation} (improved by looping, since iterating refines representations) and knowledge \emph{capacity} (improved by memory, since unique parameters store more information): ``the core tradeoff is between knowledge manipulation, which looping enhances as it repeatedly refines the representations, and knowledge capacity, which requires additional unique parameters.''

The works above converge with ours from very different regimes. \citet{yu2026mesh} arrive at memory-based externalization as a fix for diagnosed pathologies in LM recursion; \citet{frey2026adaptive} arrive at memory-plus-looping as a complementary architecture combining math and commonsense performance; \citet{geiping2025recurrent} and \citet{zhu2025ouro} demonstrate the scaling viability of depth recurrence at LM scale; we arrive at memory tokens as the minimum architectural element required for functional behavior in a minimal UT on combinatorial reasoning. Different scales, different tasks, different memory mechanisms---the same architectural conclusion in each case: the residual stream alone is insufficient for sustained recursive computation, and some form of explicit state externalization is needed.

\subsection{Adjacent: External Memory and Adaptive Long Context}

For completeness, we note two adjacent lines of work that share vocabulary with ours but address different problems. Memformer \citep{wu2020memformer}, Infini-attention \citep{munkhdalai2024infini}, and Titans \citep{behrouz2025titans} address long-context efficiency through external memory modules updated across sequence segments. The Neural Turing Machine \citep{graves2014ntm} and Differentiable Neural Computer \citep{graves2016dnc} introduced explicit differentiable memory with content and location-based addressing. These target sequence-length recurrence or explicit data-structure operations rather than depth recurrence within a single forward pass, but inform the design-space discussion below.

\subsection{Mapping the Design Space}

Architectures across the threads above can be characterized along three axes: (i) how the recurring unit is structured, (ii) how persistent state is carried across recursion, and (iii) how computation is allocated. Table~\ref{tab:designspace} maps the closest neighbors. Our work occupies the simplest configuration along each axis: a single transformer layer as recurring unit, sample-independent recursion-updated memory tokens as persistent state, ACT-mediated variable depth as computation allocation, and a single forward pass without deep supervision.

\begin{table}[t]
\centering
\caption{Architectural comparison across recursive-transformer variants. UTM (this work) occupies the simplest configuration along each axis.}
\label{tab:designspace}
\scriptsize
\setlength{\tabcolsep}{3pt}
\renewcommand{\arraystretch}{1.15}
\resizebox{\textwidth}{!}{%
\begin{tabular}{@{}lp{2.0cm}p{2.4cm}p{2.4cm}p{2.2cm}p{2.2cm}p{2.4cm}p{1.8cm}@{}}
\toprule
Property & UTM (ours) & HRM & TRM & URM & MeSH & Frey et al. & Ouro \\
\midrule
Domain & Sudoku-Extreme & Sudoku, Maze, ARC & Sudoku, Maze, ARC & Sudoku, ARC & LM (Pile) & LM (FineWeb-Edu) & LM (7.7T tokens) \\
Recurring unit & 1 transformer layer & Two networks (4 layers each) at two timescales & 2 layers, single network (Att or MLP-Mixer) & Multi-layer with ConvSwiGLU & Shared core with iteration-specific routers & 12 transformer layers, each loops independently & Shared layer stack \\
Persistent state & Concatenated memory tokens (sample-independent, recursion-updated) & Hidden states $z_L$, $z_H$ summed into input & Latents $y$ (proposed solution) and $z$ (latent reasoning), summed into input embedding & None explicit beyond hidden state & Multi-slot memory buffer & Per-layer + global key-value memory, gated & None (hidden state only) \\
Memory shape & $N \times D$ (small set) & full sequence $\times D$ (each) & full sequence $\times D$ (each) & --- & full-sequence-shaped slots & $M_L \times D$ per layer + $M_G \times D$ global & --- \\
Halting & Graves cumulative ACT & Q-learning ACT (halt + continue) & Q-learning ACT (halt-only) & Graves-style ACT (token-level) & None (fixed $N_\mathrm{loop}$) & PonderNet-style geometric & PonderNet-linked exit gate \\
Training procedure & Single forward pass & Deep supervision ($\le$16 steps) + 1-step gradient & Deep supervision ($\le$16 steps) + full-recursion BPTT & Truncated BPTT through Loops & Standard pretraining & Standard pretraining & Standard pretraining \\
Parameters & 3.2M & 27M & 5M (Att: 7M) & 27.3M & 160M--6.9B (1.4B headline) & ${\sim}200$M & 1.4B / 2.6B \\
\bottomrule
\end{tabular}%
}
\end{table}

The design-space view clarifies several relationships that surface vocabulary obscures.

\textbf{TRM and UTM use distinct memory mechanisms despite both involving prepended learnable tokens.} TRM's puzzle-ID embedding (\texttt{puzzle\_emb} in the public code) is concatenated to the input sequence as additional token positions, in the \citet{burtsev2020memory} lineage, but functions as a per-puzzle-ID static identifier rather than a recursion-updated scratchpad: it is set once per puzzle and re-injected unchanged at each recursion step. Our memory tokens are sample-independent (shared across all inputs of a given task) and update via attention each ponder step, functioning as a general-purpose recursion-updated scratchpad. TRM additionally maintains its full-sequence-shaped latents $y$ and $z$, summed (not concatenated) into the input embedding, as the actual recursive state---a mechanism we do not use.

\textbf{MeSH's mechanism occupies a different point in the design space than ours, despite reaching a similar conclusion.} MeSH's slots are full-sequence-shaped tensors accessed by per-iteration routers that are not weight-tied across iterations. Our memory tokens are distinct prepended positions accessed by standard self-attention with a fully weight-shared block. The two mechanisms address overlapping problems via different architectural choices: MeSH addresses functional specialization across iterations by making the routers iteration-specific; we maintain full weight-tying and address the same underlying issue (insufficient bandwidth in the residual stream) by adding parallel addressable storage. Whether iteration-specific routing and addressable storage are complementary or substitutable is an open empirical question.

\textbf{URM and UTM are closer architecturally to each other than to TRM or HRM.} Both use the explicit UT+ACT framing; both rely on the residual stream rather than per-position latent state for working memory. URM substitutes ConvSwiGLU and truncated backpropagation for our memory tokens and single-pass training. The two architectures are evaluated under different training regimes---URM follows the HRM/TRM 1K-example small-data protocol with heavy augmentation, while our 57\% is reported under full training---so a direct head-to-head comparison would be misleading; isolating the contribution of each architectural axis under a matched regime is left to future work.

\textbf{Ouro shows that loop depth alone scales without explicit memory augmentation, but identifies the same manipulation-vs-capacity tradeoff.} Ouro reports that recurrence does not increase per-parameter knowledge storage but enhances knowledge manipulation. This parallels Frey et al.'s explicit dissociation of looping (manipulation) from memory (capacity), and is consistent with our finding that memory tokens become necessary precisely when the architecture is most minimal---exactly the regime where the residual stream's capacity is most constrained.

\textbf{Our ``memory tokens are necessary'' finding is specific to the minimal configuration.} TRM solves Sudoku-Extreme without our memory-token mechanism but with full-sequence per-position summed latents and deep supervision. URM solves Sudoku-Extreme without explicit persistent state but with ConvSwiGLU and TBPTL. MeSH improves recursive language modeling without standard memory tokens but with externalized iteration-specific routing. Frey et al. recover commonsense performance via key-value memory banks. Ouro relies on loop depth alone. The finding is therefore not that memory tokens are universally necessary for recursive computation, but that \emph{something} must compensate for the residual stream's narrow bandwidth in a depth-recurrent architecture---and across the works above, every architecture that succeeds adds some such mechanism. Memory tokens are one such compensation; per-position summed latents are another; iteration-specific external buffers are a third; gated key-value retrieval is a fourth; convolutional sequence mixing is a fifth; sufficient parameter count combined with massive pretraining is a sixth. Our work establishes the minimal mechanism along this axis on the smallest architecture and cleanest task; characterizing the full equivalence class is open.

\subsection{What the Convergence Suggests}

The threads above started from different motivations---adaptive computation budgets, long-context memory, biologically-inspired recursion, parameter-efficient pretraining at scale---and have arrived in 2025--2026 at architectures that share a substantial core: weight-shared recurrence, persistent state across recursion, and (often) adaptive halting. This convergence is now visible across scales, from minimal architectures on combinatorial tasks (this work) to language models pretrained on trillions of tokens \citep{zhu2025ouro,geiping2025recurrent,yu2026mesh}.

A useful framing emerges from this convergence: looping enhances the model's ability to \emph{manipulate} information through iterative refinement, while memory enhances its \emph{capacity} to store information that the manipulation operates on. \citet{frey2026adaptive} articulate this dissociation explicitly; Ouro corroborates it independently with the empirical observation that recurrence does not increase per-parameter knowledge storage but dramatically improves multi-hop reasoning. Under this framing, our finding that memory tokens are necessary in the minimal UT setting---where residual-stream capacity is most constrained---is exactly what one would predict: the minimal residual stream cannot manipulate knowledge it has no capacity to store. The bandwidth of working memory and the depth of computation are complementary architectural resources, and the right ratio depends on task structure.

This convergence motivates several testable predictions:

\begin{enumerate}
\item \textbf{Bandwidth-allocation scaling.} Wider models (more residual capacity) should require fewer memory tokens, while deeper recursion (more state to maintain across steps) should require more---and varying these axes systematically should trace iso-performance contours rather than independent optima.
\item \textbf{Additive vs.\ substitutive gains.} Since TRM solves Sudoku-Extreme via full-sequence summed latents and URM via convolutional sequence mixing, augmenting either with explicit recursion-updated memory tokens should yield additive rather than substitutive gains; if the gains are instead substitutive, the bandwidth framing must be refined.
\item \textbf{Dissociability of halting and memory.} Within the recursive-transformer family halting and memory are dissociable mechanisms---Ouro and PonderNet provide adaptive halting without explicit memory augmentation, while MeSH provides explicit memory without adaptive halting (fixed $N_\mathrm{loop}$)---so we expect that combining them deliberately (rather than treating them as solutions to the same problem) is the architecturally correct move.
\end{enumerate}

Characterizing the full equivalence class of bandwidth-augmenting mechanisms across tasks and scales, and verifying or falsifying these predictions, is the natural next direction for the field.

\section{Future Work}

Our findings open several directions we consider most promising: (1) \textbf{Scaling and width revisited}: our 3.2M model is half the size of TRM (7M). A hidden=768 model (${\sim}7$M params) would enable direct comparison. Notably, all prior width experiments (hidden=768) used the default initialization and uniformly failed (2--5\% EM)---it is unknown whether deep-start initialization would rescue these, which would reframe the ``width-variance trap'' observed in early experiments as another manifestation of the router initialization issue. (2) \textbf{Muon optimizer}: Newton-Schulz-based optimizers \citep{jordan2024muon} have shown promise for weight-shared architectures; combining Muon with deep-start initialization and lambda warmup is unexplored. (3) \textbf{TRM-style training protocol}: our small-dataset results (6--8\% generalization) use a single-pass encoder; adapting TRM's iterative answer-refinement mechanism to the UT framework could improve algorithmic generalization. (4) \textbf{Multi-task evaluation}: testing on maze navigation, ARC-AGI, or formal logic to determine whether the $T{=}8$ threshold and attention specialization patterns are task-universal or Sudoku-specific. (5) \textbf{Deeper attention analysis}: per-head, per-difficulty probing to understand whether memory readers and constraint propagators emerge consistently across tasks and scales.

\section{Limitations}

(1) \textbf{Single task}: all results on Sudoku-Extreme; thresholds are likely task-specific. (2) \textbf{Architecture-specific}: HRM, TRM, and URM all solve Sudoku without memory tokens, via different mechanisms (hierarchical $z_L,z_H$ state; per-position $y,z$ latents with deep supervision; ConvSwiGLU + TBPTL, respectively); our findings apply to the single-block UT. A parameter-matched comparison (7M) is planned. (3) \textbf{Algorithmic generalization}: the single-block UT memorizes but does not generalize (6--8\% vs TRM's 87.4\%) on the 1K-example protocol, likely reflecting architectural differences in learning mechanisms rather than capacity alone (3.2M vs TRM's 7M params). (4) \textbf{$T{=}64$ seed sensitivity}: deep-start eliminates sensitivity in the plateau but not at the dilution boundary. (5) \textbf{Fixed-depth variance}: fixed-depth processing shows high seed variance ($\pm 9.3\%$ across 3 seeds) compared to ACT ($\pm 0.7\%$), likely due to reliance on a single step's representation. (6) \textbf{Evaluation granularity}: eval EM is computed on 12.8K-sample batches (SE $\pm$0.4pp), well below the inter-seed variance ($\pm$0.7pp); full 423K test-set evaluation would not materially change the reported means.

\section{Conclusion}

We demonstrate that learned memory tokens are empirically necessary for a single-block Universal Transformer to solve combinatorial reasoning tasks. Without them, no configuration succeeds---regardless of initialization, seed, ponder steps, or use of ACT. With 8 or more tokens (${\sim}1$ per 10 puzzle cells), the model reliably achieves 57\% exact-match on Sudoku-Extreme, with a stable plateau through $T{=}32$.

A key methodological finding enables these results: in our experiments, standard ACT initialization creates a degenerate $p \approx 0.5$ equilibrium causing $>$70\% of runs to fail. This trap---confirmed inherent to ACT via normalization ablation---is fixable with a single line of code.

ACT provides more consistent results than fixed-depth processing ($56.9 \pm 0.7\%$ vs $53.4 \pm 9.3\%$), and lambda warmup achieves matching accuracy ($57.0 \pm 1.1\%$) using 34\% fewer ponder steps---genuine compute efficiency without quality loss. Across the $T{=}8$--$32$ plateau, memory tokens and ponder depth substitute at fixed accuracy. Under halt-side pressure ($\lambda{=}0.001$+warmup, $S{=}0$), the halt-vs-$T$ curve is monotonically decreasing ($11.60 \to 11.50 \to 10.29 \to 8.25$ for $T \in \{8, 16, 32, 64\}$) at near-constant accuracy on the plateau, with a 2pp drop at the dilution boundary ($T{=}64$). This is the canonical depth-state trade-off: memory tokens and recursive depth function as substitutable resources for combinatorial reasoning, with the dilution boundary intrinsic to the architecture rather than an artifact of saturated halting.


\appendix
\section{Puzzle Solving Visualization}

\begin{figure}[h]
\centering
\includegraphics[width=\textwidth]{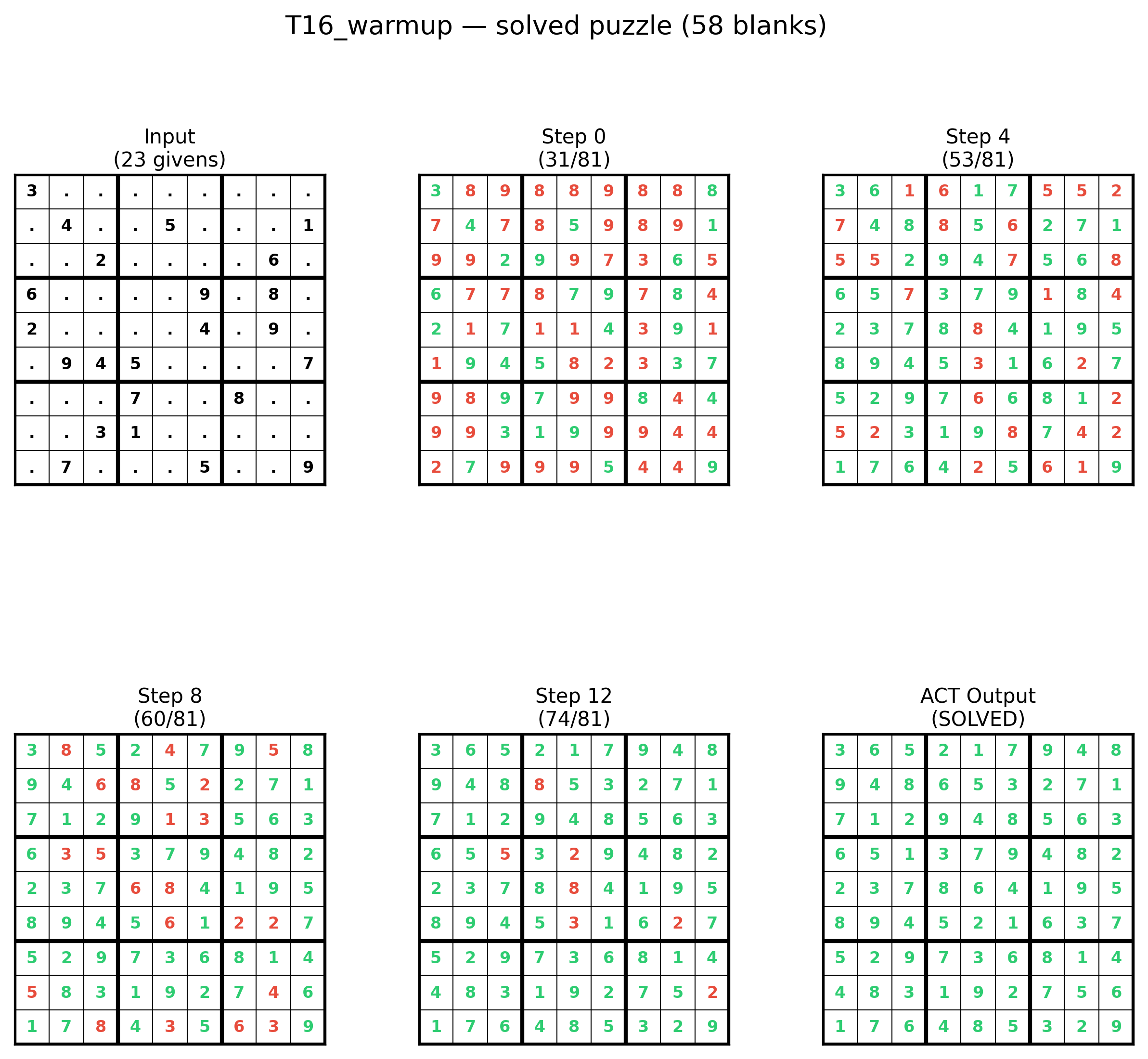}
\caption{Step-by-step puzzle solving ($T{=}16$, $\lambda{=}0.001$+warmup). The model progressively fills cells across ponder steps: 31/81 $\to$ 53 $\to$ 60 $\to$ 74 $\to$ SOLVED. Green = correct, red = error.}
\label{fig:puzzle_solved}
\end{figure}

\begin{figure}[h]
\centering
\includegraphics[width=\textwidth]{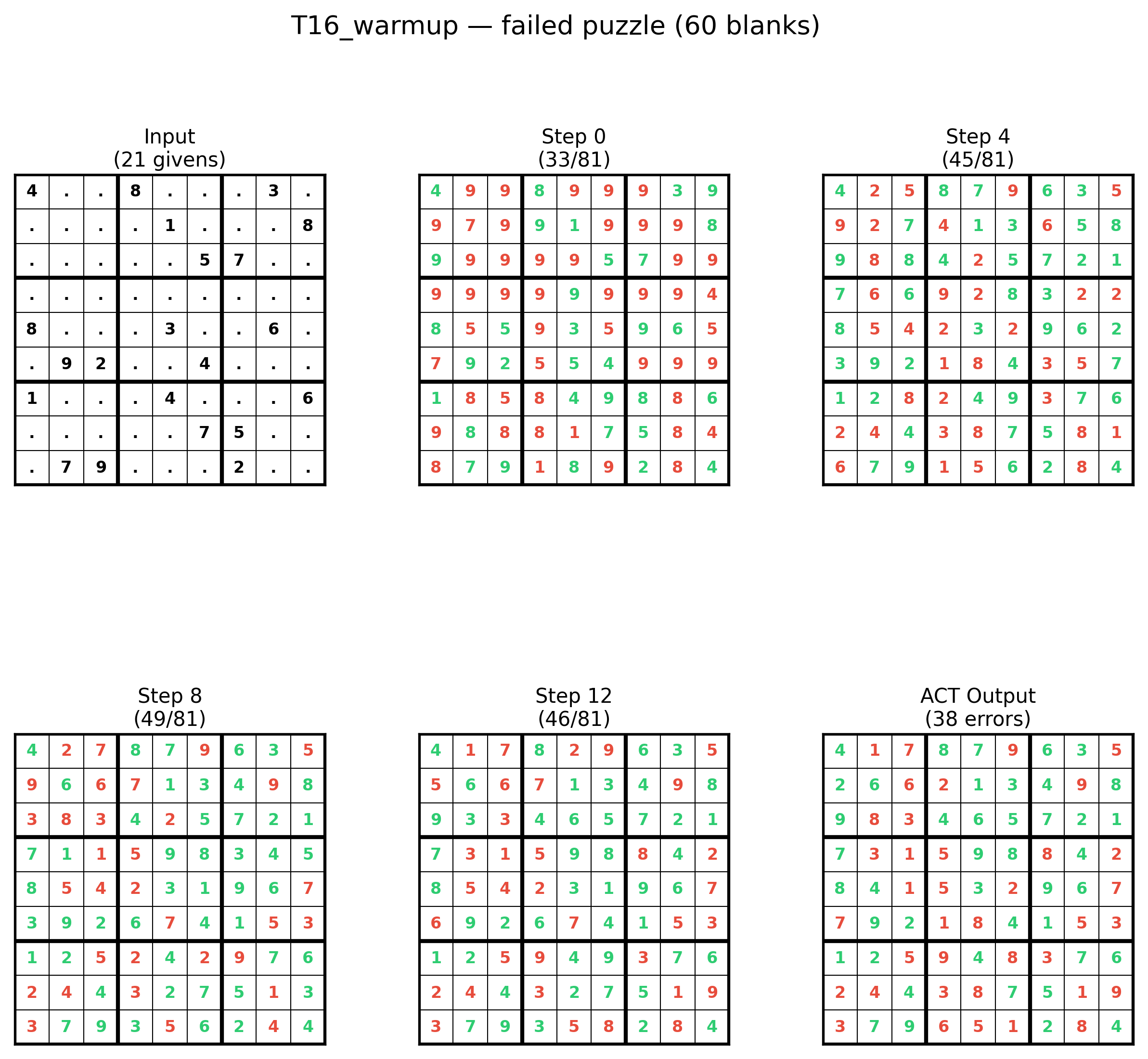}
\caption{A puzzle the model fails to solve. Despite 18 ponder steps, the model reaches only 43/81 correct cells (38 errors), with errors concentrated in regions requiring deep constraint propagation.}
\label{fig:puzzle_failed}
\end{figure}

\end{document}